\documentclass[letterpaper, 10 pt, conference]{ieeeconf}  %

\IEEEoverridecommandlockouts                              %

\usepackage[english]{babel}

\usepackage{graphics} %
\usepackage{epsfig} %
\usepackage{mathptmx} %
\usepackage{times} %
\usepackage{amsmath}
\usepackage{multirow}
\usepackage{booktabs} %
\usepackage{amssymb}  %
\usepackage{comment}
\usepackage{subfigure}
\usepackage{url}

\title{\LARGE \bf
Learning Side-by-Side Navigation for Personal Mobility Vehicles from Human Drivers
}

\author{Luis Yoichi Morales and Florent Ferreri
  \thanks{*yoichims@ieee.org}}

\newcommand{\argmax}{\arg\!\max}

\newcommand{\resApartObsCcwDrivers}{17.2}
\newcommand{\resApartObsCcwOurs}{2.1}
\newcommand{\resApartObsCwDrivers}{17.8}
\newcommand{\resApartObsCwOurs}{1.3}

\newcommand{\resBackupsLayCcw}{2.8}
\newcommand{\resBackupsLayCw}{0.1}
\newcommand{\resBackupsObsCcw}{0.2}
\newcommand{\resBackupsObsCw}{2.2}

\newcommand{\resConvFreeCcwDrivers}{52.9}
\newcommand{\resConvFreeCcwOurs}{81.2}
\newcommand{\resConvFreeCwDrivers}{56.0}
\newcommand{\resConvFreeCwOurs}{91.5}
\newcommand{\resConvObsCcwDrivers}{34.0}
\newcommand{\resConvObsCcwOurs}{60.5}
\newcommand{\resConvObsCwDrivers}{36.0}
\newcommand{\resConvObsCwOurs}{68.9}

\newcommand{\resConvRunsObsCcw}{6}

\newcommand{\resLatMeanFree}{1.16}

\newcommand{\resLatSdFree}{0.16}

\newcommand{\resModeCloseAlong}{0.68}
\newcommand{\resModeCloseConv}{29}
\newcommand{\resModeCloseDur}{6}
\newcommand{\resModeCloseLat}{0.65}
\newcommand{\resModeCloseSep}{1.00}
\newcommand{\resModeWideAlong}{0.54}
\newcommand{\resModeWideConv}{56}
\newcommand{\resModeWideDur}{4}
\newcommand{\resModeWideLat}{1.41}
\newcommand{\resModeWideSep}{1.59}
\newcommand{\resRunsConditions}{24}

\newcommand{\resRunsLayouts}{24}

\newcommand{\resSpeedFree}{0.89}
\newcommand{\resSpeedObs}{0.87}

\newcommand{\resStoppedObsCwOurs}{1.5}

\begin{document}

\maketitle
\thispagestyle{empty}
\pagestyle{empty}

\begin{abstract}

  This paper presents an approach to achieve side-by-side autonomous motion
  with a pair of personal mobility vehicles (PMVs). Given a global path, two
  PMVs plan cooperatively so as to hold a social formation that lets the
  passengers hold a conversation, while keeping the ride safe and smooth.
  Each vehicle runs its own sampling-based model-predictive planner; the
  planners exchange
  nothing but the odometry each vehicle could observe of the other, and the
  joint decision is expressed as a pair of utility matrices over pairs of
  candidate trajectories, resolved by a common rule.
  The behavioural parameters are fitted by inverse reinforcement learning from
  $409$ minutes of driving by ten pairs of people in human-driven PMVs in an
  indoor environment, and the lanes the pair can use are derived offline from
  the corridor itself rather than fixed by hand. A safety layer the cost cannot
  override executes a move only if the vehicle can still brake clear, measured
  from its whole body, and a vehicle that is stuck reverses along its own path
  and continues. In closed-loop simulation over
  $\resRunsConditions$ runs in the conditions the drivers were recorded in and
  $\resRunsLayouts$ in four generated obstacle layouts, no vehicle entered a
  wall or a cone. The pair holds conversational formation for
  $\resConvObsCcwOurs\%$ and $\resConvObsCwOurs\%$ of the time among obstacles
  (anticlockwise and clockwise) and $\resConvFreeCcwOurs\%$ and
  $\resConvFreeCwOurs\%$ in a clear aisle, against the drivers'
  $\resConvObsCcwDrivers\%$ and $\resConvObsCwDrivers\%$, and
  $\resConvFreeCcwDrivers\%$ and $\resConvFreeCwDrivers\%$, and it matches
  their speed. It keeps
  one lateral formation where the drivers alternated between a close diagonal
  and a wide abreast one, and it varies far less.
  Experimental results are based on a differential drive kinematic simulator running on ROS~2\footnote{The source code is available at \url{https://github.com/yoichims/side_by_side_ros2}.}.

\end{abstract}

\section{Introduction}
\label{1.introduction}
Current robot systems can achieve reliable long-term navigation in dynamic environments with pedestrians around \cite{Morales_jfr09, kummerle2014autonomous}. Recent works have placed attention to the pedestrian flow to be able to pass-by walking people in a social manner \cite{predicting_irl, mehta2016iros, DBLP:journals/corr/ChenELH17}. These robot navigation approaches consider the humans passing by and the overall safety but do not consider their own motion smoothness.

Differently from standard mobile robots, autonomous PMV's carry human passengers, so the navigation must be safe and comfortable for the passenger.
PMV's can provide transportation services in a smaller scale than car-like vehicles in locations such as hospitals, retirement homes, houses, attraction parks. Moreover, in some remote rural areas, there is the need to provide mobility assistance to isolated aged people where PMV's have the potential to provide personal last mile services such as autonomus ride shares \cite{shared_autonomous_vehicles} and autonomous personal vehicles on demand \cite{autonomy_for_mobility_on_demand}.

While PMV's could provide mobility services in remote areas and independence for aged people \cite{senior_mobility_needs}, it is important to allow the passengers to move together in groups with partners allowing them to engage in conversations with other pedestrians or PMV passengers.
Mobility independence and the ability of socializing are relevant factors that impact well-being and health \cite{senior_mobility_needs},
impacting the quality of life of aged persons as leaving home allows people to keep social and physical activities preventing mental capacity decline on aged people \cite{socially_integrated_dementia, social_relation_decline}.
Moreover, it has been reported that poor or limited social interactions increases the risk of dementia \cite{social_network_on_dementia}.

Regarding people interacting in groups, it has been found that the majority of people tend to move in groups %
\cite{journal.pone.0010047}.
At low densities people tend to walk in small social groups of two to five people where they tend to walk in side-by-side formation in order to keep direct line of sight to engage in interaction \cite{PhysRevE.89.012811}.
Within large groups subgroups composed of two or three people emerge \cite{zanlungo2013walking}.
People act cooperatively with their partners, keep social formation and anticipate their movement to keep pace.
Despite being previous works and proposed approaches to develop models to keep pace with their side by side partners \cite{garrell2012cooperative, kuno_2014, repiso2017line, 4115bf2699a34801ae8484d50c76b0d4},
currently there are no %
autonomous navigation systems which can reliably move alongside partners in a cooperative way or be integrated into a social group.

Humans achieve side-by-side motion by performing shared planning. %
In a shared planning activity humans share goals and intentions and hold the partner accountable for his role \cite{bratman2013shared}. While moving together as a shared activity we shared the common intention of moving towards a destination which is different from the case of moving alongside a stranger by chance where we just happen to collaborate or negotiate traffic for a short period of time \cite{bratman2013shared}.
\begin{figure}[tb]
 \centering
 \includegraphics[width=3in, angle=0]{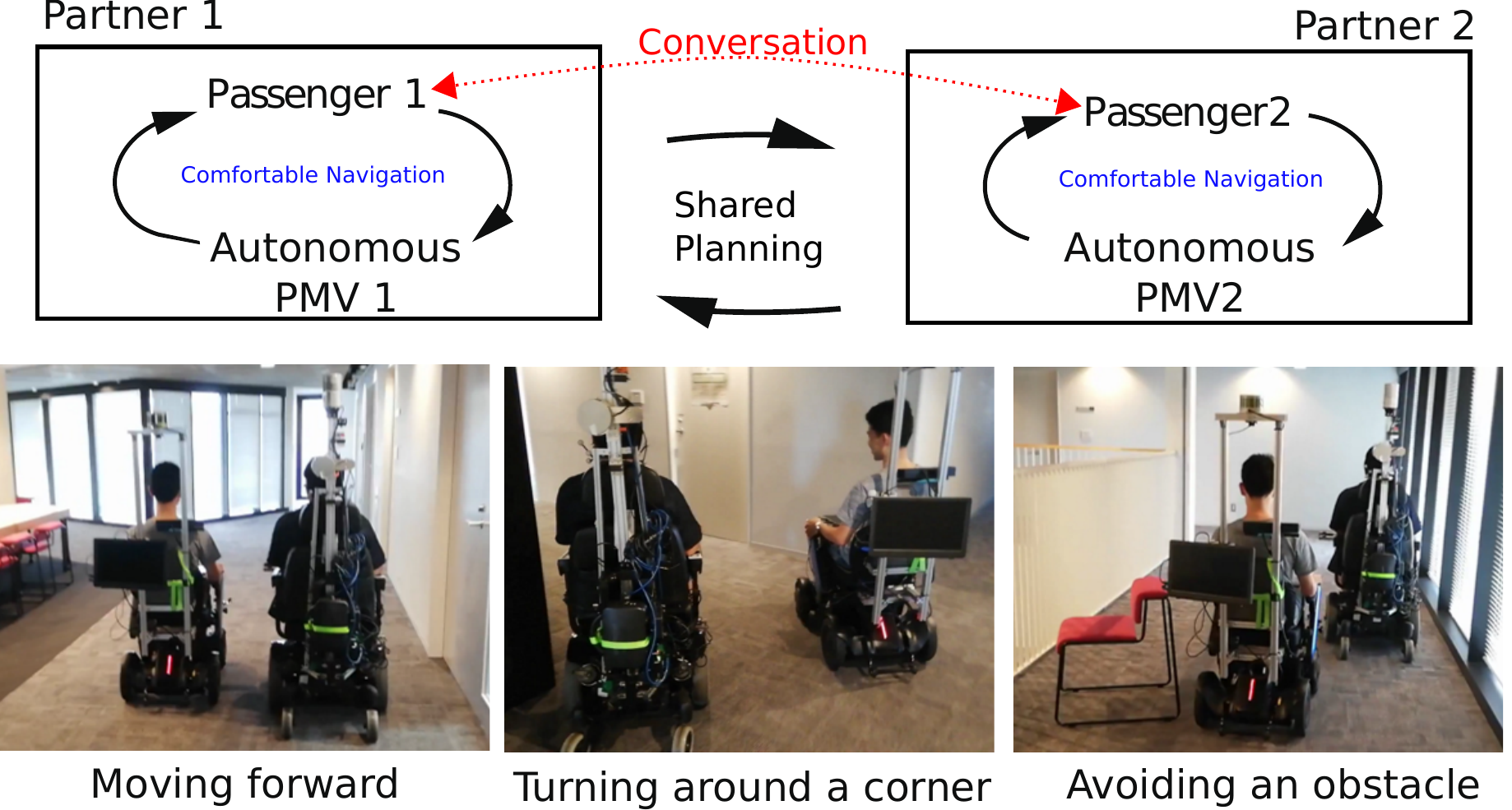}
   \caption{ Two personal mobility vehicles performing shared planning to move side by side enabling the passengers to engage in a conversation. }
\label{fig:concept}
\end{figure}
Figure \ref{fig:concept} shows an image of a pair of %
partners moving alongside as a social mobile group. An agent is composed of a passenger and a PMV. The PMV
transports the passengers while he/she has a conversation with another partner.
The vehicles share their pose and velocities, anticipate the partner's future location and jointly plan their next control action to be able to keep side-by-side formation. A motion planner computes the linear and angular velocities of each vehicle so that the navigation results in smooth and comfortable for each passenger. %

In our knowledge, this is the first work which treats autonomous navigation of PMV's as a social group where anticipation of the partner for side-by-side formation is crucial to keep a social interaction between passengers.
The contributions of this work are:

\begin{itemize}

\item \emph{A decentralised formulation of the joint decision.} The pair's
  choice is expressed as a pair of utility matrices over pairs of candidate
  trajectories, scored by each vehicle independently.
  The matrices they construct are transposes of one another.
  Nothing about intention is communicated: the only quantity crossing between the vehicles is odometry, which
  keeps the formulation valid when communication is absent.

\item \emph{Formation geometry derived from pairs of human drivers.}
  Whether two vehicles fit side by side at all, are computed offline from the surveyed
  map, per sample of the reference path %
  which makes the method portable to new environments.

\end{itemize}

\section{Related Works}
\label{sec:related}

Previous works have studied the human phenomenon of moving together.
A landmark work regarding pedestrian behavior is the social force model where behavior is modeled as
several forces actuating within the pedestrians \cite{social_force_model}. Several extensions to the model have been proposed,
for example \cite{Zanlungo_2015} models pedestrian group dynamics as a microscopic potential orthogonal to the direction of movement.
The work in \cite{journal.pone.0010047} studied the impact of social groups in crowds where it was found that at low density people tend to move side-by-side. %
An approach to model pedestrian dynamics in a social group formed of two pedestrians was proposed in \cite{PhysRevE.89.012811}. %
The work in \cite{Costa2010} presented a study on proxemic behaviors in groups from dyads to five-person groups. %
Force model approaches need parameter calibration which is usually environment dependent. Moreover, placement of the goal locations for the agents
is not trivial and it could be context dependent. Subgoal computation requires large ammounts of pedestrian data to compute \cite{zanlungo_rss}.
Work regarding walking together as a social activity was discussed in \cite{Gilbert_1990} where the concepts of shared action and plural subject as a group of two or more persons that have a common goal were introduced.
In navigation state of the art, it is common to give a global path so that the robot can follow \cite{dijkstra_59} %
which can be customized to consider human factors in their cost function \cite{MORALES201813}.
In the work of this paper, we model a group composed of two PMV's moving alongside as a shared activity where a global path is given in advance.

A model based on a shared planning approach where each walking partner anticipates the future location of their partner and jointly plans their future location was proposed in \cite{morales_JHRI_14}.
An extension to this model was presented in \cite{murakami2014destination}, where the pair could hold the formation even when one agent did not know the goal, by switching between a leader--follower and a collaborative state according to how reliably the follower could estimate the route.
Additionally, the work in \cite{saiki2017social} presented a model for three robots moving as a social group replicating pedestrian behavior.
Regarding robot companions, an approach of cooperative robots accompaning people was presented in \cite{garrell2012cooperative} and a model that allows a robot to approach and engage a group formed of two pedestrians walking side-by-side was presented in \cite{repiso2018robot}.

Work on safe and smooth navigation for PMV's has been done on single robotic wheelchair path planning for smooth autonomous navigation was presented in \cite{Gulati_09, Park_12, MORALES201813}.  Regarding robotic wheelchairs there are previous approaches that dealt with the task of moving with companions \cite{kuno_2014}. The closest work to our proposal is the work presented in \cite{multiple_wheelchair_communication, maintaining_formation} where they introduce a wheelchair system that can keep a formation for users to to be able to avoid obstacles and communicate. These works reactively compute the linear and angular velocities of their own motions.

Differently from related works, in our approach, the motion is between two agents that carry passengers maximizing mutual utility rather than a robot and a human.
The two planners reach compatible decisions from their own observations to keep a formation so that passengers can converse.

\section{Problem Definition}
This section presents the definition of PMV's, a definition of social mobility and provides the problem statement.%

\subsection{Personal Mobility Vehicles Moving as Social Groups}
A personal mobility vehicle is a self-contained independent vehicle that transports a single passenger. %
PMV's navigate on pedestrian paths
interacting with pedestrians and are not regulated by car-like traffic rules.

In pedestrian behavior research, a social group consists of two or more individuals who interact socially while moving toward a common goal
\cite{pedestrian_behavior_1982},
\cite{Gilbert_1990, tomasello2009we}.
Interaction is the oral communication usually accompanied with non-verbal communication such as gestures and gaze exchange \cite{judithHall}.
When walking together, people tend to move side-by-side to keep direct line of sight while communicating \cite{Kleinke1986}.

\subsection{Shared Planning}
\label{sub:shared_planning}

Shared planning is a joint intentional activity in which two or more agents
cooperate to achieve a joint goal \cite{bratman2013shared}. A group of PMVs
engaged in shared planning has the common goal of following a path to a
destination. They navigate cooperatively, mutually opening space and keeping a
social formation that allows the passengers to converse.
Each vehicle runs its own sampling-based model-predictive planner at
Every cycle it samples its own candidate trajectories, samples a prediction of the partner's
from the partner's \emph{observed} state, scores every pair of candidates,
selects one pair by a solution concept, and executes only the first command of
its own half of that pair.

\subsection{Problem Definition}
\label{sec:definition}
In this work we consider a group composed of two vehicles and a pre-defined global path. We center on the ego-vehicle ``e'' ($^e$) who cooperatively navigates with a partner ``p'' ($^p$).
The ego-vehicle plan its path considering the utility of his own movement and the partner.
Both vehicles run a sampling-based model-predictive planner with a time horizon $T$.
The system interaction state is given by:
\begin{eqnarray}
{\bf x}=( {\bf x}^e, {\bf x}^p, m, \rho )
\end{eqnarray}
where $m$ is the environment map and $\rho$ is a global path.

The ego-vehicle state is given by
\begin{eqnarray}
  {\bf x}^e_{t+1}=f^e({\bf x}^e_t, u^e_t, m, \rho)
\end{eqnarray}
where $u_t$ is the control. The partner vehicle state is

\begin{eqnarray}
{\bf x}^p_{t+1}=f^p({\bf x}^p_t, u^p_t, m, \rho)
\end{eqnarray}

The interaction dynamic system state is given by
\begin{eqnarray}
{\bf x}_{t+1}=f({\bf x}_t, u^e_t, u^p_t).
\end{eqnarray}

The optimal action for the ego-vehicle ${{\bf u}^e}^*$ is given by:
\begin{eqnarray}
  \begin{split}
    {{{\bf u}^e}}^* = \argmax_{{\bf u}^e}  \{  U^e( {\bf x}_t, {\bf u}^e, {\bf u}^p, m, \rho) (1-\gamma)  \\
    + \; \; U^p ( {\bf x}_t, {\bf u}^p, {\bf u}^e, m, \rho) \gamma \}
  \end{split}
\end{eqnarray}

where %
$\gamma \in [0,1]$ is a parameter of ``consideration'' for the partner's utility of motion. %
This variable models social or moral status preferences such as giving a partner the way \cite{hall2005nonverbal}, \cite[p.~135-149]{knapp2013nonverbal}.
The ego-vehicle only considers itself with a value of $\gamma = 0$ and only considers the partner with $\gamma =1$. %
We considered both partners equally and set the value to $\gamma=0.5$ for the rest of this work.

\subsection{Vehicle Model and Candidate Generation}
\label{sec:model}
The PMV accepts speed and yaw-rate commands and follows them through a
first-order actuation lag,
\begin{equation}
  \dot{v} = \frac{v_c - v}{\tau_v}, \qquad
  \dot{\omega} = \frac{\omega_c - \omega}{\tau_\omega},
  \label{eq:lag}
\end{equation}
with $\tau_v = 0.41$~s and $\tau_\omega = 0.49$~s identified from the corpus.

\subsection{The Joint Decision Over Candidate Pairs}
\label{sec:game}
Let $\{\mathbf{u}^e_i\}_{i=1}^{n}$ be the ego vehicle's candidates and
$\{\mathbf{u}^p_j\}_{j=1}^{m}$ the candidates it predicts for its partner.
Scoring every pair yields two matrices,
\begin{equation}
  A_{ij} = U^e(\mathbf{u}^e_i, \mathbf{u}^p_j), \qquad
  B_{ij} = U^p(\mathbf{u}^e_i, \mathbf{u}^p_j),
  \label{eq:bimatrix}
\end{equation}
where $U^e$ and $U^p$ are the same functions evaluated with the two roles
exchanged. Because every feature is computed from the pair rather than from
one vehicle's privileged view, the matrices the partner constructs satisfy
$A^p = (B^e)^{\!\top}$ and $B^p = (A^e)^{\!\top}$. The two vehicles are
therefore solving one problem seen from two sides, which is what allows a
cooperative concept to be evaluated independently by each vehicle without
either transmitting its choice.

\section {Utility Function Features}
\label{sec:utility}

The utility of a candidate pair is a weighted sum of feature deviations,
\begin{equation}
  U^j = -\sum_{x \in \mathcal{F}} \frac{w_x}{\sigma_x} \,
        \big\langle \lvert f_x - f^\star_x(\mathbf{c}) \rvert \big\rangle ,
  \qquad j \in \{e, p\},
  \label{eq:utility}
\end{equation}
where $\langle\cdot\rangle$ averages over the steps of the horizon.
$f_x$ is the feature, $f^\star_x(\mathbf{c})$ its preferred value in the
current context $\mathbf{c}$, $\sigma_x$ a scale making a unit weight mean
equal influence, and $w_x$ the weight. Each term is a deviation from a preferred value $f^\star_x$.
The preferred value depends on the context $\mathbf{c}$, one set of
weights describes both a pair riding abreast and in single file (when avoiding obstacles).
The weights never change, only the values they are measured against.

The context is the set of observed quantities the preferred values key on,
\begin{equation}
  \mathbf{c} = \big(\kappa,\; o,\; s,\; \varsigma\big),
  \label{eq:context}
\end{equation}
where $\kappa$ is the signed curvature of the reference path at the pair's
midpoint, $o$ the distance from the pair to the nearest detected object, and
$s$ the arc-length position along the route. The last term,
$\varsigma \in \{-1, 0, +1\}$, says whether this vehicle is on the
\emph{outside} or \emph{inside} of a curve or neither.

\subsection{Features}
\label{sub:features}

The feature set $\mathcal{F}$ is grouped by what each term relates to:
\emph{relative} terms couple the two vehicles, \emph{motion} terms constrain
one vehicle's own comfort, and \emph{path} terms tie the pair to the route.
Table~\ref{tab:design} lists every term with its target, scale, and the
weight fitted and used; where each of $f^\star_x$, $\sigma_x$ and $w_x$ comes from is set out
in \S\ref{sec:fitted}.

Which terms have a non-zero preferred value is what gives the deviation form
its content. For acceleration, range rate and distance to one's own lane the
preferred value is zero, and asking for a small deviation is the same as
asking for a small quantity. For the terms that carry the formation it is
not: the drivers held a separation of about $1.4$~m, a speed set by the
curvature of the path, and a bearing near abreast, none of which is an
extremum of anything. %

Two conventions matter. The formation angle follows the definition in our
earlier work, $R_a = \mathrm{ang}(p_i) - \mathrm{ang}_i(p_j, p_i)$, so that
$R_a = \pi/2$ is abreast; the cost uses $\sin R_a$, which is invariant to the
sign convention since $\sin(\pi - \theta) = \sin\theta$. And progress is the
only reward in the set, entering negated, which is what keeps the pair moving
when every other term is satisfied by standing still.

Four relative terms --- range rate, relative yaw rate, relative yaw and the
along-track offset --- carry zero weight in the shipping configuration.
Separation and $R_a$ are polar coordinates of the same displacement, so
$\{d_R, R_a\}$ already spans both degrees of freedom of the formation and the
along-track offset is implied rather than independent.

\subsection{Fitting the Weights by Inverse Reinforcement Learning}
\label{sub:irl}
Of the three fitted objects (\S\ref{sec:fitted}), only the weights
$w_x$ are obtained here. The planner samples a finite candidate set, so the
action space of the imitation problem is discrete. We use maximum-entropy inverse
reinforcement learning to find the weights of each feature. We model a driver at sample
$t$ as choosing candidate $i$ from the generated set $\mathcal{U}_t$ with

\begin{equation}
  P(i \mid \mathbf{w}) =
    \frac{\exp\!\left(-\mathbf{w}^{\!\top}\mathbf{f}_i\right)}
         {\sum_{j \in \mathcal{U}_t}
          \exp\!\left(-\mathbf{w}^{\!\top}\mathbf{f}_j\right)} ,
  \label{eq:maxent}
\end{equation}

where $\mathbf{f}_i$ is the scale-normalised feature vector of candidate $i$
and the demonstrated choice $i^{\star}_t$ is the candidate whose rollout comes
closest to the motion the driver actually executed over the same horizon.
Fitting maximises the regularised mean log-likelihood

\begin{equation}
  \mathcal{L}(\mathbf{w}) = \frac{1}{T}\sum_{t=1}^{T}
    \log P(i^{\star}_t \mid \mathbf{w})
    - \lambda \lVert \mathbf{w} \rVert^{2} ,
  \qquad \mathbf{w} \ge \mathbf{0} ,
  \label{eq:irlobj}
\end{equation}

with $\lambda = 10^{-3}$, where larger weights describe a more deterministic driver.

\section{Data Collection and Fitting}
\label{sec:experimental}

\subsection{Robot System}
\begin{figure}[tb]
 \centering
 \includegraphics[width=3.2in, angle=0]{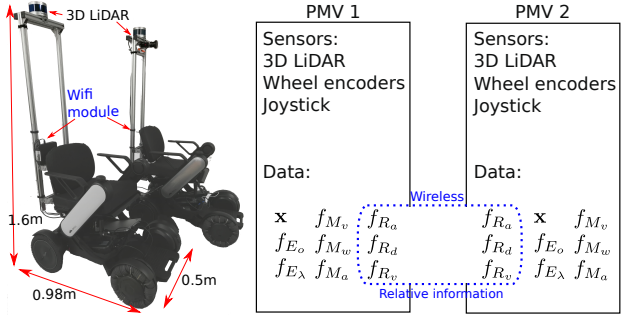}
 \caption{ PMV's system comfiguration. Each vehicle is equiped with a $3D$ LiDAR and wheel encoders. Vehicle parameters are shared through wifi.}
\label{fig:whill}
\end{figure}

For data collection and experimentation we used two personal mobility vehicles (WHILL Model C) each equipped with a LiDAR (Velodyne VLP-16) and wheel encoders (see Fig. \ref{fig:whill}).
Each robot localized towards a previously built map and shared its pose, linear velocity and angular velocity wirelessly to the partner. Data of both vehicles was time stamped and recorded simultaneously.

\subsection{Data Collection}
\label{sub:data}

Twenty people, forming ten pairs, drove the PMV's and were paid for their participation in the experiment.
Drivers within a pair had the same degree of seniority to avoid social status biases while driving together (it has been reported that status differences tend to lead to lower interpersonal distances \cite{hall2005nonverbal}, \cite[p.~135-149]{knapp2013nonverbal}.
Additionally, participant pairs were required to know themselves beforehand so that they could have a natural conversation interaction.

\begin{figure}[tb]
 \centering
 \includegraphics[width=3.5in, angle=0]{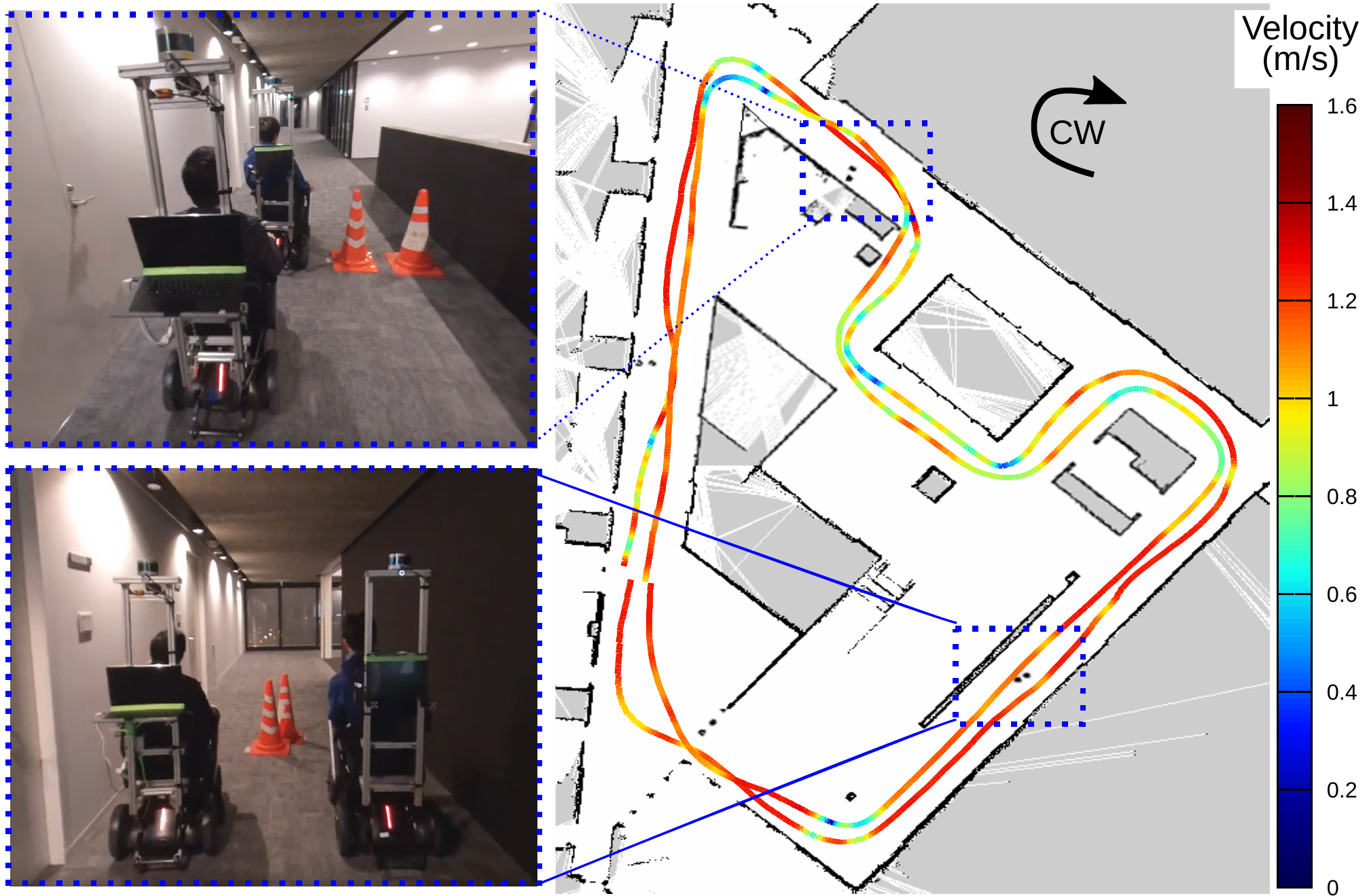}
 \caption{ %
   Trajectories of PMV's moving CW while the drivers were engaged in a conversation in an indoor environment. The color represents the velocity of each vehicle. Some cones were placed as obstacle in half of the experimental conditions.}
\label{fig:env}
\end{figure}

The experiment was performed in an indoor office environment shown in Fig. \ref{fig:env}. Environmental corridor widths were between $2.24$ and $2.60$ $m.$.
First, we walked participant pairs through the experimental course and instructed them where to drive.
We allowed participants to sit down and have practice runs during a period of $10~minutes$. We then asked them to drive while they had a conversation while following the instructed path for $10~minutes$.
Driver pairs were instructed to drive side-by-side as they were asked to have a conversation.
For data collection, participants drove a total of $24$ laps in sets of $3$ laps.
After each set participants stopped briefly. $4$ sets were driven free from obstacles ($T_F=\{T_{F_{cw}}, T_{F_{ccw}}\} $) and $4$ sets were driven with obstacles ($T_O=\{T_{O_{cw}}, T_{O_{ccw}} \} $).
$2$ sets were driven clockwise (CW) and $2$ sets counterclockwise (CCW), for each CW and CCW condition, participants switched driving sides.

The corpus comprises all $10$ pairs ($20$ drivers). Each pair drove the
loop in both directions, in a clear aisle and with cones placed in the
corridor as obstacles, so every pair contributes to all four conditions.

\begin{figure*}[t]
  \centering
  \includegraphics[width=\linewidth]{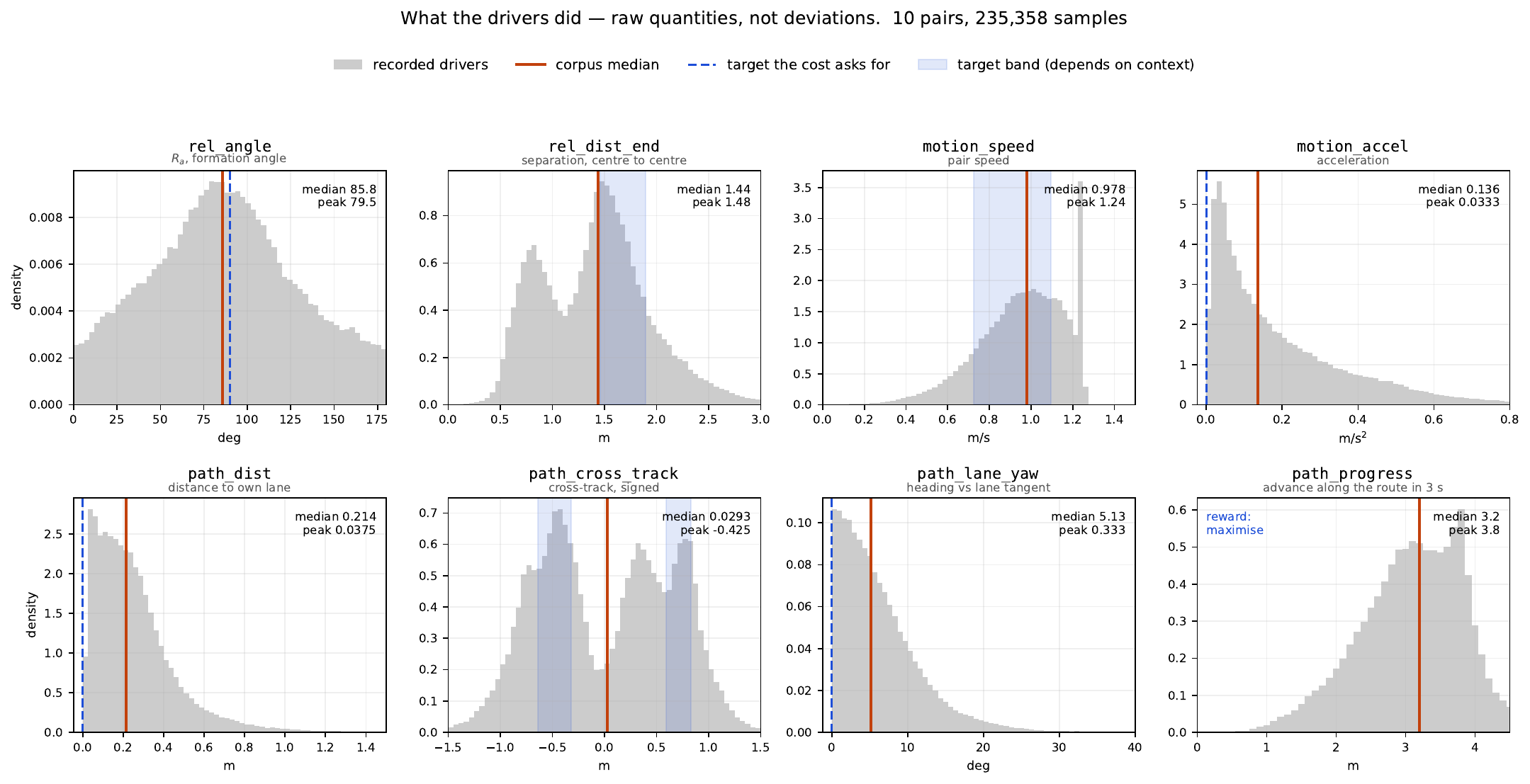}
  \caption{What the drivers did: one panel per term of
    Table~\ref{tab:design}, the raw quantity pooled over the ten recorded
    pairs, with the corpus median in orange and the target the cost asks for
    dashed --- a shaded band where that target is a function of context rather
    than a constant. For the signed cross-track position the target is the
    vehicle's lane offset, so the two bands are the two lanes; progress is the
    only reward and has no target. %
    $R_a$ follows the definition of our earlier side-by-side work,
    $R_a = \mathrm{ang}(p_i) - \mathrm{ang}_i(p_j, p_i)$, and peaks just under
    $90^\circ$. Two distributions carry their own warnings.
    \texttt{motion\_speed} has a hard spike at $1.24$~m/s, which is the
    vehicle's governor and not a preference, so the drivers' speed is
    censored from above. \texttt{rel\_dist\_end} is \emph{bimodal}, at
    $0.85$ and $1.50$~m, because the pair switches between riding abreast and
    filing, and the two formations hold different distances; the split is
    within pairs, not between them. Fitting a unimodal target to it averages
    across two behaviours and lands between them.}
  \label{fig:raw_hists}
\end{figure*}

In total the corpus holds $85$ recordings and $409$ minutes ($6.8$~h) of
paired driving, close to evenly divided between conditions and directions
(Table~\ref{tab:corpus}): $201$ minutes in the clear aisle and $208$ with
obstacles, $198$ CCW and $211$ CW. A recording lasts a median of $281$~s
($140$--$529$~s), and each pair contributes $8$--$10$ recordings and
$32$--$51$ minutes. Both vehicles' poses and velocities were recorded at
$17$--$20$~Hz, resampled to a common $10$~Hz, and split at gaps longer than
$0.5$~s, discarding pieces shorter than $5$~s; this yields $94$ continuous
segments and $245\,608$ paired samples. One pair's recordings carried a
clock running at three times real time and were rescaled before use.
Figure~\ref{fig:raw_hists} shows the resulting distributions of every
quantity the cost is built on.

\begin{table}[t]
  \caption{The human-driven corpus: recordings and minutes of paired
    driving per condition and direction, $10$ pairs.}
  \label{tab:corpus}
  \centering \footnotesize
  \begin{tabular}{lccccc}
\toprule
& \multicolumn{2}{c}{anticlockwise} & \multicolumn{2}{c}{clockwise} & total \\
\cmidrule(lr){2-3}\cmidrule(lr){4-5}
condition & runs & min & runs & min & runs \\
\midrule
clear aisle & 21 & 99 & 21 & 102 & 42 \\
surveyed cones & 21 & 100 & 22 & 108 & 43 \\
\midrule
\textbf{total} & \textbf{42} & 198 & \textbf{43} & 211 & \textbf{85} \\
\bottomrule
\end{tabular}

\end{table}

\begin{table}[t]
  \caption{The utility terms the planner uses.
  }
  \label{tab:design}
  \centering \footnotesize
  \begin{tabular}{@{}lllc@{}}
\toprule
term & measured as & $\sigma$ & weight \\
\midrule
bearing $R_a$ (signed) & $\langle|\sin R_a - s|\rangle$ & 0.133 & 0.53 \tabularnewline  %
separation \emph{(social distance)} & $|d_T - d^\star_R(o)|$ & 0.3\,m$^\ddagger$ & 2.00 \tabularnewline  %
\midrule
speed & $\langle|v - v^\star(\kappa)|\rangle$ & 0.2\,m/s$^\ddagger$ & 3.00 \tabularnewline  %
acceleration & $\langle|\dot v|\rangle$ & 0.2\,m/s$^2$ & 1.05 \tabularnewline  %
\midrule
path distance & $\langle\mathrm{dist}(\cdot, \mathrm{lane})\rangle$ & 0.3\,m$^\ddagger$ & 1.91 \tabularnewline  %
cross-track (signed) & $|\langle c\rangle - c^\star|$ & 0.203\,m & 6.00 \tabularnewline  %
lane yaw & $\langle|\psi - \psi_{\mathrm{lane}}|\rangle$ & 0.092\,rad & 3.00 \tabularnewline  %
progress & $-(s_T - s_0)$ & 2\,m & 0.95 \tabularnewline  %
\bottomrule
\end{tabular}

\end{table}

\subsection{Parameter Fitting}
\label{sec:fitted}
The weights $w_x$ of features of section \ref{sub:features} are constants, fitted by maximum-entropy inverse reinforcement learning using the human driven data of section \ref{sub:data}.

Each demonstration is one recorded instant. From the recorded state of one
vehicle, the planner's own rollout generator produces the candidate set the
planner would choose from, and every candidate is scored with the features of
Table~\ref{tab:design}. The relative features are evaluated against the
partner's \emph{recorded} future rather than a prediction of it, so that the
weights describe the drivers' preferences instead of absorbing the errors of
a partner model. The candidate whose rollout comes closest to what the driver
actually did over the same $3$~s is taken as the driver's choice $i^\star_t$,
and the weights maximise the regularised log-likelihood (\ref{eq:irlobj}) of
those choices under the model (\ref{eq:maxent}), with $\lambda = 10^{-3}$ and
$\mathbf{w} \geq \mathbf{0}$.

Both vehicles of a pair are demonstrations of the same cost, one on the outer
lane and one on the inner, so each sampled instant contributes up to two
samples. We draw $900$ samples from the ten pairs, $25\%$ of them from
instants below $0.6$~m/s. The drivers were almost always at cruise, but the
planner starts from rest and slows for obstacles, and a fit that never sees
low-speed states leaves standing still unpriced.

Validation holds out whole driver pairs, since the question is whether the
preferences generalise across people: the model is refitted ten times, each
time without one pair, and scored on that pair's choices. Held out this way,
the driver's actual choice ranks at $0.898$ on average among the candidates
(normalised, $1$ for the best and $0.5$ for chance), between $0.848$ and
$0.934$ over the ten pairs, with a mean log-likelihood of $-3.360$. On the
full set the fitted weights reach $-3.346$, against $-3.786$ for equal
weights and $-4.774$ for chance. The fit therefore improves on uniform
weighting, and the improvement holds on pairs it has not seen.

A fit that ranks candidates at a single instant cannot see effects that build
up over a run, such as the pair swapping sides or spinning in place, so
several of the weights the planner runs differ from the fitted ones.

\subsection{Safety and Recovery}
\label{sec:safety_method}
The fitted cost decides how the pair moves; whether a vehicle may move at all
is decided by hard rules on top of it, which no weight can trade away.

\emph{Body clearance.} Every check measures the distance from the vehicle's
footprint rectangle, not from its centre, to the walls of the map and to the
detected objects. The candidate obstacle points are taken around the whole
outline of the body: taking only those nearest the centre misses a second
wall at a corner, and was measured to report clearance for a body already
inside a wall.

\emph{Never execute a blocked move.} A candidate is executed only if its
first $0.5$~s keeps the body clear of walls and objects and if, after
executing it, the vehicle can still brake to rest without contact. The
braking test integrates the stop through the actuation lag and assumes the
command acts for one planning tick plus the measured delay between the state
it was computed from and its execution, so a planner that falls behind
becomes more cautious rather than late. When no candidate passes, the vehicle
stops; it never takes the least-bad candidate.

\emph{Recovery.} A vehicle that has been unable to move for half a second
reverses along the path it has itself just driven, which is known to be
free, under the same braking test, and then rejoins at reduced speed. If it
stops again at the same place, it reverses twice as far. Objects are
remembered for $30$~s after they were last seen, because the laser does not
cover the region behind the vehicle.

\emph{Staying together.} The vehicle ahead along the route stops when its
partner falls more than $3.0$~m behind, and resumes when it is within
$2.2$~m, the $99$th and $90$th percentiles of the drivers' separation among
cones. It starts such a hold only when no object is within $2.5$~m, so that
it never waits in a gap its partner still has to pass through.

\section{Experimental Procedure and Results}
\label{sec:experiment}
The planner was evaluated in closed loop in a simulation of the surveyed
seventh-floor environment, with the occupancy map taken from the same survey
used for the data collection and the objects entered as data rather than baked
into the map. Both travel directions were run, since the route is not
symmetric: the two directions differ in where the corridor pinches and in
which lane an object blocks. Each run lasted $170$~s, long enough for roughly
one circuit. The two human conditions were run with $\resConvRunsObsCcw$
repeats per direction each ($\resRunsConditions$ runs) and the generated
layouts with three ($\resRunsLayouts$ runs), all with one frozen version of the
planner and a metrics node sampling both vehicles every tick.

\subsection{Three Evaluation Conditions}
\label{sec:layouts}
The planner is evaluated in three conditions, which support different kinds of
claim and should not be read as one table.

\emph{Clear aisle} and \emph{surveyed cones} reproduce the two conditions the
drivers were recorded in: the same route with nothing in it, and the same
route with the cones the drivers actually met, entered as data at their
surveyed positions. Because a human counterpart exists for both, these support
claims about \emph{fidelity} --- whether the pair moves the way the drivers
moved --- and Sections~\ref{sec:conversation}
and~\ref{sec:distance} are measured in them. The corpus keeps the two apart,
and we do too: the drivers held $1.09$~m of lateral separation in a clear
corridor and closed to $0.99$~m at an obstacle, so pooling the conditions mixes
a bimodal population into one distribution and makes both modes look like
error.

\emph{Generated layouts} is a third condition with no human counterpart. The
lane offsets the planner follows are derived offline, so a test against the cones they were derived from cannot
distinguish knowledge that generalises from knowledge that was fitted. We
therefore generate cone
layouts along the route: placements are sampled on the centreline rather than
uniformly in the plane, so every cone sits somewhere the pair will actually
meet it, each is placed by kind (on the outer lane, the inner lane, the
centre, or against an edge) rather than at a uniform random offset, and each
is checked to leave beside it a free band at least as wide as the tightest
site in the surveyed environment. Cones are written as circles at the measured
radius. Nobody ever drove these layouts, so they support claims about
\emph{generalisation} and safety --- contact, stalls, progress, and the
fraction of the route spent in file rather than side by side --- but never
about resemblance to human behaviour.

Two limits apply to this third condition throughout. Four layouts with three
repeats is enough to establish mechanisms and to compare arms on stalls and
progress, but not to estimate rates: separating a $10\%$ stall rate from a
$5\%$ one needs of the order of $74$ runs per arm. And contact depth at a
given site varies more between sweeps of an identical configuration than
between the arms we compare, so we report sites and mechanisms rather than
depths.

\subsection{Conversation and Formation}
\label{sec:conversation}
Table~\ref{tab:conversation} divides every moment of the runs by what the pair
was doing, beside the drivers in the same condition. The quantity the system
exists for is conversational formation: the partner within $25^\circ$ of
abreast and less than $2$~m away. In all four combinations of condition and
direction the pair holds it for more of the time than the drivers did:
$\resConvObsCcwOurs\%$ and $\resConvObsCwOurs\%$ among the surveyed cones,
against the drivers' $\resConvObsCcwDrivers\%$ and $\resConvObsCwDrivers\%$,
and $\resConvFreeCcwOurs\%$ and $\resConvFreeCwOurs\%$ in the clear aisle,
against $\resConvFreeCcwDrivers\%$ and $\resConvFreeCwDrivers\%$.

The difference comes mostly from the pair not separating. Among the cones the
drivers spent $\resApartObsCcwDrivers\%$ and $\resApartObsCwDrivers\%$ of the
time $2$~m or more apart, the pair $\resApartObsCcwOurs\%$ and
$\resApartObsCwOurs\%$; filing past the cones took a comparable share for both.
The one combination in which a vehicle spent noticeable time stopped is
clockwise with the cones, $\resStoppedObsCwOurs\%$ of the time, which is the
recovery reported in Section~\ref{sec:safety}.

\begin{table*}[t]
  \caption{Share of time by what the pair was doing, drivers (drv.) against
    the planner (ours), in the two conditions the drivers were recorded in.
    Side by side: the partner within $25^\circ$ of abreast and under $2$~m
    away.}
  \label{tab:conversation}
  \centering \footnotesize
  \begin{tabular}{@{}lcccccccc@{}}
\toprule
& \multicolumn{4}{c}{surveyed cones} & \multicolumn{4}{c}{clear aisle} \\
\cmidrule(lr){2-5}\cmidrule(lr){6-9}
& \multicolumn{2}{c}{ccw} & \multicolumn{2}{c}{cw} & \multicolumn{2}{c}{ccw} & \multicolumn{2}{c}{cw} \\
\% of time & drv. & ours & drv. & ours & drv. & ours & drv. & ours \\
\midrule
side by side & 34.0 & \textbf{60.5} & 36.0 & \textbf{68.9} & 52.9 & \textbf{81.2} & 56.0 & \textbf{91.5} \\
filing & 48.0 & 37.3 & 45.8 & 28.4 & 37.1 & 18.8 & 36.1 & 7.3 \\
apart $\geq 2$\,m & 17.2 & 2.1 & 17.8 & 1.3 & 9.2 & 0.0 & 7.0 & 0.3 \\
one stopped or reversing & 0.8 & 0.1 & 0.4 & 1.5 & 0.8 & 0.0 & 0.9 & 0.9 \\
\bottomrule
\end{tabular}

\end{table*}

\subsection{How Far From the Drivers?}
\label{sec:distance}
Shares of time say what the pair does but not whether it moves like a human
pair, and the natural measure --- the distance between our distribution and the
drivers' --- has no meaningful scale on its own. We therefore report it against
the distance between the drivers themselves: for each recorded pair we compute
the Wasserstein distance from that pair's distribution to the pool of all the
others, which is how far a human pair typically sits from human behaviour at
large. A model closer to the pool than a typical driver pair is, on that
quantity, not distinguishable from one.

\begin{table*}[t]
  \caption{Wasserstein distance from our distribution to the pooled drivers,
    against the spread among the drivers themselves: the median distance from
    one recorded pair to the pool of all the others ($10$ pairs per direction
    and condition).}
  \label{table:distance}
  \centering \footnotesize
  \begin{tabular}{llcccc}
\toprule
& & \multicolumn{2}{c}{\textit{simulator / drivers}} & \multicolumn{2}{c}{Wasserstein distance} \\
\cmidrule(lr){3-4}\cmidrule(lr){5-6}
dir & quantity & mean & s.d. & sim vs.\ drivers & driver vs.\ drivers \\
\midrule
\multicolumn{6}{l}{\emph{clear aisle}} \\
ccw & lateral sep. & 1.17 / 1.09 & 0.14 / 0.45 & 0.28 & 0.08 \\
ccw & along-track & 0.35 / 0.59 & 0.29 / 0.55 & 0.24 & 0.13 \\
ccw & speed & 0.87 / 0.91 & 0.24 / 0.25 & 0.05 & 0.11 \\
cw & lateral sep. & 1.15 / 1.01 & 0.18 / 0.44 & 0.27 & 0.09 \\
cw & along-track & 0.29 / 0.63 & 0.32 / 0.56 & 0.34 & 0.17 \\
cw & speed & 0.92 / 0.91 & 0.30 / 0.26 & 0.06 & 0.10 \\
\midrule
\multicolumn{6}{l}{\emph{surveyed cones}} \\
ccw & lateral sep. & 1.01 / 0.99 & 0.36 / 0.48 & 0.20 & 0.13 \\
ccw & along-track & 0.63 / 0.97 & 0.53 / 0.73 & 0.34 & 0.09 \\
ccw & speed & 0.87 / 0.93 & 0.24 / 0.24 & 0.06 & 0.08 \\
cw & lateral sep. & 1.03 / 0.95 & 0.32 / 0.48 & 0.21 & 0.08 \\
cw & along-track & 0.53 / 0.99 & 0.48 / 0.72 & 0.46 & 0.27 \\
cw & speed & 0.86 / 0.93 & 0.30 / 0.26 & 0.07 & 0.08 \\
\bottomrule
\end{tabular}

\end{table*}

\begin{figure*}[t]
  \centering
  \includegraphics[width=0.85\textwidth]{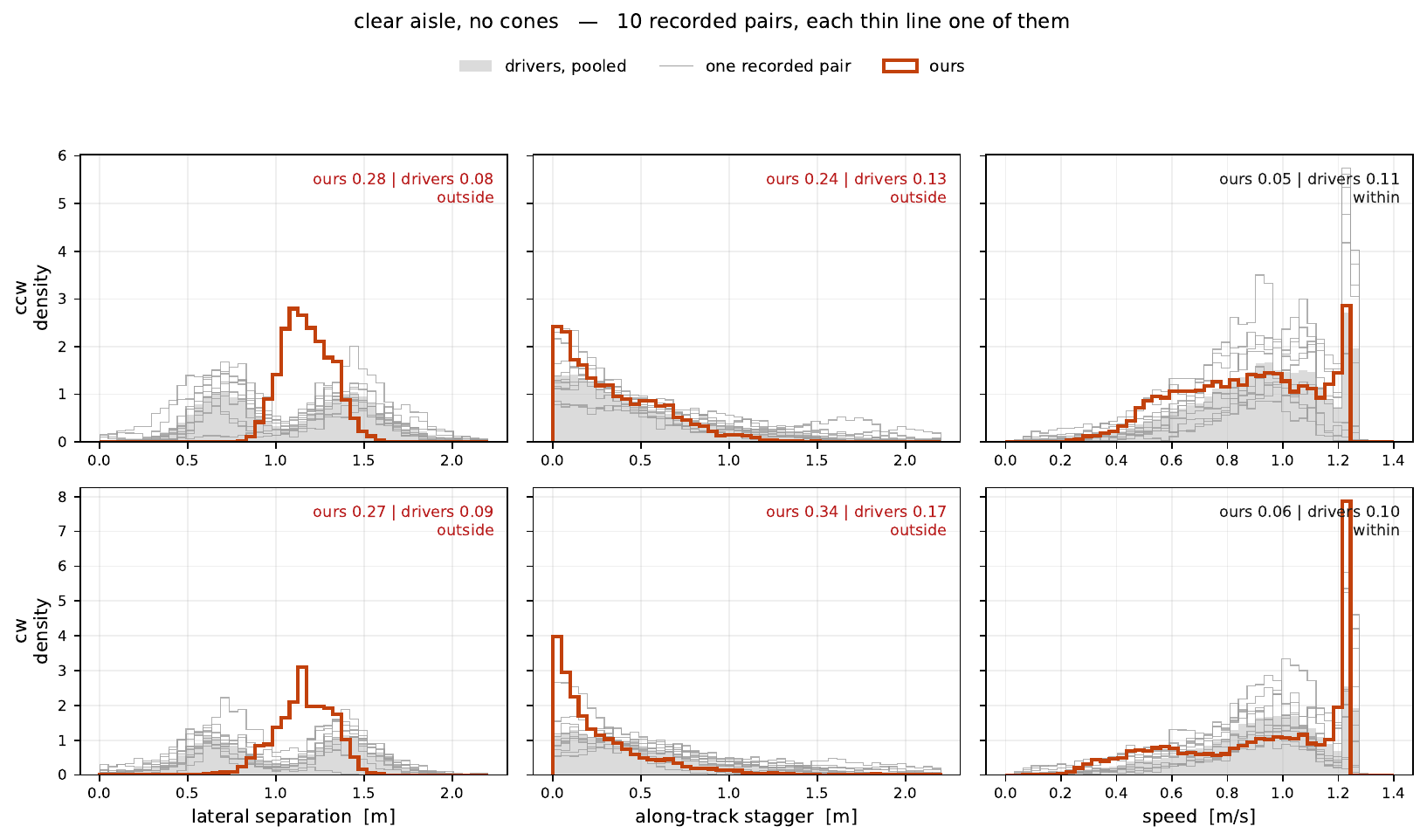}
  \caption{Clear aisle. Pooled driver behaviour filled, each recorded pair a
    thin line, ours over the top. A curve inside the spread of the thin lines
    is not distinguishable from a human pair on that quantity.}
  \label{fig:dist_free}
\end{figure*}

\begin{figure*}[t]
  \centering
  \includegraphics[width=0.85\textwidth]{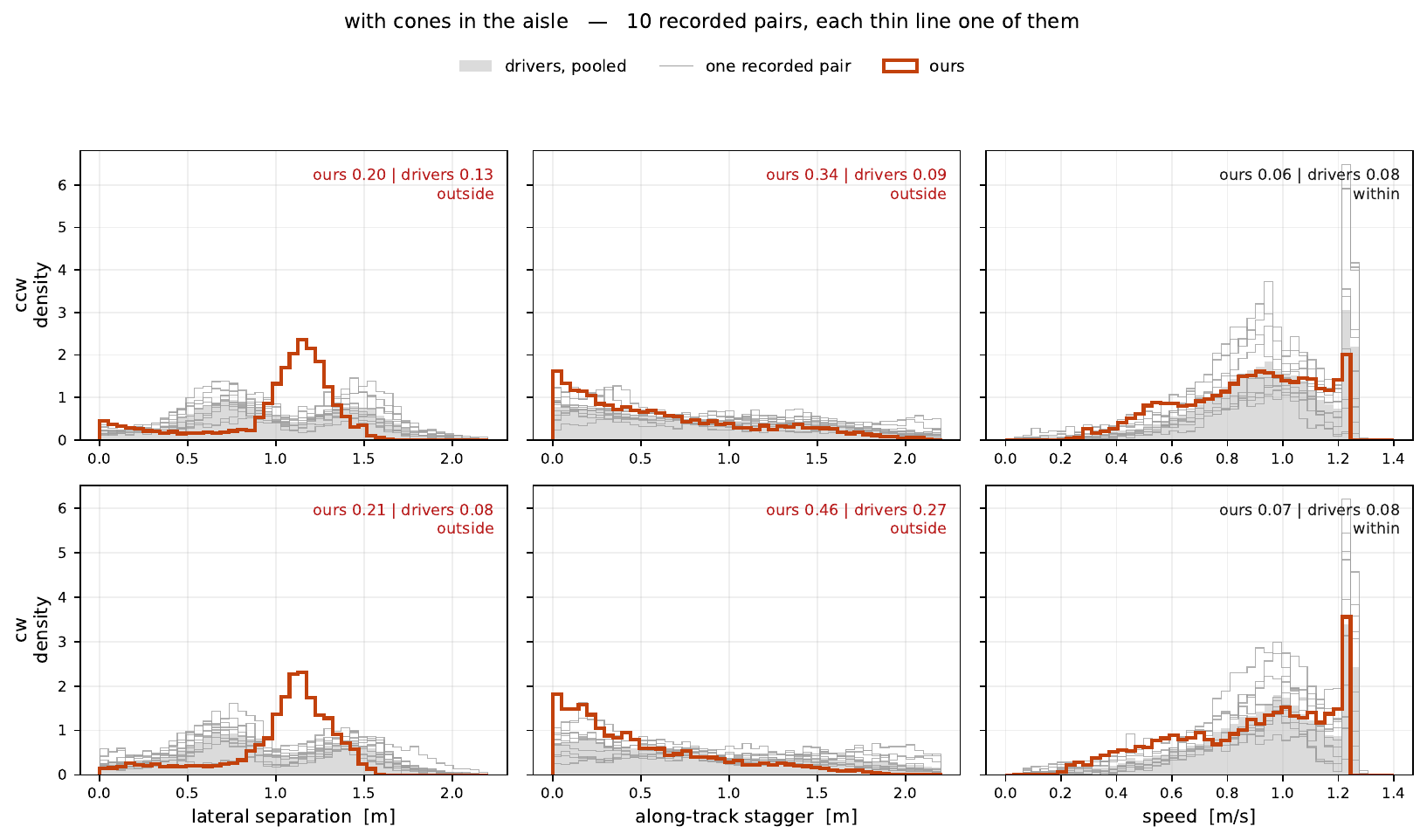}
  \caption{Surveyed cones, same construction.}
  \label{fig:dist_obstacle}
\end{figure*}

Speed is the one quantity on which the pair is human-like: in both
conditions and both directions it lies inside the spread of the drivers. Its
mean is $\resSpeedFree$~m/s in the clear aisle and $\resSpeedObs$~m/s among
the cones, against the drivers' $0.91$ and $0.93$~m/s, and its speed
distribution peaks where theirs does, at $1.24$~m/s, the speed at which the
vehicles' governor capped them. The planner's maximum speed is set to that
governor, and its preferred straight-line speed too: fitted to the drivers'
median speeds it comes out at $1.09$~m/s, biased low because the drivers'
speed is censored at the governor. Among the cones the pair is slightly
slower than the drivers, because the braking test of
Section~\ref{sec:safety_method} slows it wherever the room to stop is short.
On the other eight comparisons of
Table~\ref{table:distance} the pair is outside the spread of the drivers, and
the differences are systematic rather than noisy. The pair rides more level
than the drivers, with less stagger along the route. It is also far more
uniform: in the clear aisle the
standard deviation of its lateral separation is $\resLatSdFree$~m, against
$0.44$--$0.45$~m for the drivers.

The largest difference in shape is the lateral separation, which is bimodal
for the drivers and unimodal for the pair. The two human modes are two
formations. In the \emph{close} one (lateral separation under $1$~m) the pair
rides staggered on a diagonal --- medians $\resModeCloseLat$~m across,
$\resModeCloseAlong$~m along the route and $\resModeCloseSep$~m apart --- and
in the \emph{wide} one it rides abreast, $\resModeWideLat$~m across,
$\resModeWideAlong$~m along and $\resModeWideSep$~m apart. They are not a
property of the corridor: the position along the route explains less than a
tenth of which mode a pair is in, every one of the ten pairs spends between
$38$ and $60\%$ of its time in the close mode, and a typical episode of either
lasts only $\resModeCloseDur$--$\resModeWideDur$~s. Each pair alternates
between the two at the same places. The cost asks for a single preferred
separation, so the pair holds one separation of about $\resLatMeanFree$~m
between the two modes, and it cannot reproduce the alternation without a
second formation and a rule for switching between them, which the data do not
supply. For the purpose of the system the difference is not a loss: only
$\resModeCloseConv\%$ of the drivers' close-mode time is conversational,
against $\resModeWideConv\%$ of the wide mode, so a pair that keeps the abreast
formation instead of drifting into the diagonal is holding the better of the
two for conversation --- which is also why it exceeds the drivers in
Table~\ref{tab:conversation}. It is therefore not indistinguishable from a
human pair: it is a more conservative and more consistent one, which holds the
formation the conversation needs for more of the time.

\subsection{Safety and Liveness}
\label{sec:safety}
Table~\ref{tab:safety} reports contact, liveness and recovery in all three
conditions. Wall contact is measured independently of the planner, as the
deepest penetration of the body into any surveyed map cell, and counted above
half a map cell ($25$~mm).

In the two conditions the drivers were recorded in, $\resRunsConditions$ runs,
no part of either vehicle entered a wall or a cone and no vehicle ended a run
stuck. Recovery was needed mostly among the surveyed cones, where a vehicle
backed up $\resBackupsObsCcw$ times per run anticlockwise and
$\resBackupsObsCw$ clockwise, and resumed every time; in the clear aisle a
vehicle backed up once in twelve runs. The generated layouts are harder than
anything the drivers met. Over $\resRunsLayouts$ runs no part of either
vehicle entered a wall or a cone, and the vehicles backed up
$\resBackupsLayCcw$ times per run anticlockwise and $\resBackupsLayCw$
clockwise. Two anticlockwise runs ended with a vehicle stopped, counted as
stuck in Table~\ref{tab:safety}. Neither vehicle was stuck: each was holding
for its partner (Section~\ref{sec:safety_method}), which was still reversing
its way out of a blocked corner $3.5$--$5.7$~m behind when the run ended.

\begin{table*}[t]
  \caption{Contact, liveness and recovery. Contact is measured from the
    logged poses, independently of the planner, against every surveyed map
    cell and every placed cone; a run counts when the body entered a wall or a
    cone by more than $25$~mm, and \emph{worst} is the deepest penetration.
    Stuck: runs ending with a vehicle stopped.}
  \label{tab:safety}
  \centering \footnotesize
  \begin{tabular}{@{}llrrrrrrrrr@{}}
\toprule
condition & dir & runs & \multicolumn{2}{c}{wall contact} & \multicolumn{2}{c}{cone contact} & stuck & backups & holds & stopped \\
\cmidrule(lr){4-5}\cmidrule(lr){6-7}
& & & runs & worst & runs & worst & runs & /run & & s/run \\
& & & & mm & & mm & & & & \\
\midrule
surveyed cones & ccw & 6 & 0 & 0 & 0 & 0 & 0 & 0.2 & 1 & 0 \\
 & cw & 6 & 0 & 0 & 0 & 0 & 0 & 2.2 & 0 & 6 \\
\midrule
clear aisle & ccw & 6 & 0 & 0 & --- & --- & 0 & 0.0 & 0 & 0 \\
 & cw & 6 & 0 & 0 & --- & --- & 0 & 0.2 & 1 & 2 \\
\midrule
generated layouts & ccw & 12 & 0 & 0 & 0 & 0 & 2 & 2.8 & 6 & 26 \\
 & cw & 12 & 0 & 0 & 0 & 0 & 0 & 0.1 & 1 & 1 \\
\bottomrule
\end{tabular}

\end{table*}

\begin{table}[ht]
  \caption{Open-loop agreement with the drivers: distance between the chosen
    rollout's endpoint and the drivers' realised position 3~s later, over 300
    recorded timesteps.}
  \label{table:concepts_open}
  \begin{center}
  \begin{tabular}{|l|c|c|c|}
    \hline
    concept & mean & median & vs.\ oracle \\
    \hline
    oracle (action-space floor) & 1.058 & 0.256 & --- \\
    \hline
    optimistic  & 1.603 & 1.120 & $+0.544$ \\
    social      & 1.645 & 1.145 & $+0.587$ \\
    maximin     & 1.663 & 1.159 & $+0.605$ \\
    anticipated & 1.757 & 1.307 & $+0.699$ \\
    \hline
    random (chance) & 2.014 & 1.705 & $+0.956$ \\
    independent     & 2.301 & 2.109 & $+1.242$ \\
    \hline
  \end{tabular}
  \end{center}
\end{table}

The result that separates cleanly is not among the cooperative concepts: it is
\emph{independent}, which ignores the partner and is worse than choosing a
feasible rollout at random. Reasoning about the partner at all is what the
open-loop comparison establishes. The four concepts that do reason about the
partner span $0.15$~m, a sixth of the gap between the best of them and chance.
Their end-of-horizon separations are more informative than their errors: the
drivers ended $1.439$~m apart, and every concept that resolves the partner's
column with a maximum overshoots that ($1.68$--$1.82$~m), while maximin, which
resolves it with a minimum, sits at $1.261$~m.

\textbf{Closed loop.} Open-loop imitation says nothing about what a concept
does when both vehicles run it simultaneously and their assumptions about each
other must be mutually consistent. We therefore ran each alternative against
the social concept over the generated layouts (Section~\ref{sec:layouts}),
four layouts in both directions with three repeats --- 48 runs per comparison
--- holding every other parameter fixed. Because the layouts differ greatly in
difficulty, the comparison is paired: each run is differenced against the
social run of the same layout, direction and repeat, so layout variance
cancels (Table~\ref{table:concepts_closed}).

\begin{table}[ht]
  \caption{Closed-loop comparison against the social concept, paired by
    layout, direction and repeat ($n = 24$ pairs of runs each). Positive is
    worse than social. Uncertainties are the standard deviation of the mean.}
  \label{table:concepts_closed}
  \begin{center}
  \begin{tabular}{|l|c|c|c|}
    \hline
    vs.\ social & cone contact & stalls/run & progress \\
    \hline
    optimistic  & $+4.5 \pm 6.3$~mm    & $-0.04 \pm 0.09$ & $-0.001 \pm 0.027$ \\
    anticipated & $-19.2 \pm 11.7$~mm  & $+0.21 \pm 0.12$ & $-0.028 \pm 0.024$ \\
    maximin     & $-4.6 \pm 3.9$~mm    & $+0.08 \pm 0.16$ & $-0.054 \pm 0.034$ \\
    \hline
  \end{tabular}
  \end{center}
\end{table}

The optimistic concept is a null on every measure: its open-loop advantage of
$0.04$~m does not survive the pair having to act on mutually incompatible
assumptions, which is what the open-loop protocol cannot test, because there
the partner's behaviour is supplied by the recording rather than generated by
the same rule. The anticipated concept costs stalls --- $+0.21$ per run, and
in the clockwise direction specifically $0.00 \to 0.42$ stalls per run with
progress falling from $0.87$ to $0.81$, in three of the four clockwise
layouts. This is the failure its structure predicts: each vehicle predicts
that the partner will take the contested gap and defers to it, and both defer.
Its apparent gain on contact is most plausibly a consequence of the same
thing, since a stopped vehicle does not strike a cone, and we do not read it
as a safety benefit.

The maximin concept, which assumes the partner does the worst thing available
rather than the best, is the only arm that loses on the task: progress falls
$0.054 \pm 0.034$ and anticlockwise stalls rise from $0.08$ to $0.42$ per run.
Assuming the worst makes the pair hesitant, which is what the concept means,
so mechanism and measurement agree. It is worth reporting for a second reason:
open loop it ends closest of all concepts to the drivers' separation
($1.261$~m against the humans' $1.439$~m), while every concept that resolves
the partner's column with a maximum overshoots it at $1.68$--$1.82$~m. The
conservative concept holds a more human distance and gets less done.

We therefore keep the social concept. It is not measurably better than the
optimistic one, but it is the only concept here under which both vehicles
provably select the same cell, and nothing in the measurements justifies
giving that guarantee up.

One limit applies to the whole of Table~\ref{table:concepts_closed}. Repeating
the identical social configuration across three separate sweeps gave $108$,
$214$ and $236$~mm as the deepest clockwise cone contact. The between-sweep
spread at that site is larger than any concept effect in the table, so these
comparisons can resolve stalls and progress but not contact depth, and we do
not claim otherwise.

The order of the repairs mattered, and is itself a result. Correcting the
floor is what made the prefix rung effective --- before it, the only candidate
the floor admitted was the fastest, which is also the one most likely to enter
the margin --- and the rung alone still left one site, which the recovery
correction closed. Over twelve runs after all three, no vehicle stalled in
either direction.

We state the strength of that evidence precisely. Twelve clean runs bound the
stall rate below roughly $25\%$ at $95\%$ confidence; against the prior rate
of $28\%$ the improvement gives a one-sided Fisher $p = 0.06$. The case rests
on the mechanisms, which are analytic, rather than on the counts. This is a
general hazard in evaluating such repairs: at a base rate near $25\%$ and
$170$~s per run, a six-run comparison cannot distinguish a fix from chance,
and two earlier changes were adopted on exactly that evidence and later
withdrawn --- one of them after it traded a stall for a vehicle grounded
$9$~mm inside a wall. We therefore verify the mechanism on recorded stall
poses replayed deterministically, and use closed-loop runs only to confirm
that nothing else regressed.

\subsection{Discussion}
\label{sub:discussion}
The pair does what the system is for: it holds conversational formation for
more of the time than the people it was fitted from, and it almost never comes
apart. It does so without resembling them closely. It matches their speed, but it keeps one lateral separation where the drivers alternated
between two, and it varies far less than they did. Two causes are clear. The separation target is a
single preferred value, so where the drivers switched every few seconds
between a close diagonal and a wide abreast formation, a cost that asks for one
value settles between the two; since the diagonal is the less conversational
of the two, we regard this as acceptable for the purpose of the system, but it
is a departure from human behaviour that a model with two formations would be
needed to remove. And the
safety rules of Section~\ref{sec:safety_method} are conservative by
construction: they slow the vehicle wherever its room to stop is short, which a
human driver, who judges risk rather than proving its absence, does not.

The safety rules carry two lessons that are not specific to this planner.
Clearance must be measured from the whole body: a check that selects its
obstacle points by distance from the vehicle's centre reported room for a body
already inside a wall at a corner, and both the planner and the measurement of
the results had relied on it. And a planner must plan from fresh state: the
odometry each planning tick used had been a quarter of a second old, and every
late stop we traced came back to it. Freshness must hold for every input, and
be checked against ground truth: a second thread that freshened the odometry
left the obstacle input queued behind the planning timer, so cones reached the
planner every few seconds instead of every scan, and the vehicles drove into
cones the planner had never been sent. Contact measured against what the
planner perceived could not reveal this; only contact measured against the
placed obstacles did.

Two issues remain open. First, at one generated layout the vehicle whose
lane is blocked does not fall in behind its partner in time, and goes round
the far side of the cones into a pocket from which it must reverse. Making it
stop and wait fixed that site but cost stalls at sites that filed well on their
own, so the decision of who passes first needs to be taken earlier, as the
drivers took it, rather than corrected late. Second, the evaluation is in simulation, with six repeats per condition and
direction, enough to establish mechanisms and large differences but not small
rates.

These point to the next step. The candidate trajectories are currently arcs in
the vehicle's own frame, which over longer horizons leave the corridor and
cannot express ``slow down, then move behind the partner''. Generating them
relative to the route --- a lateral offset that changes smoothly along it,
combined with a speed profile, as in planners for road vehicles --- would make
that manoeuvre an ordinary candidate, keep every candidate inside the corridor,
and let the horizon be extended so that the order of passing is settled early.

\section{Conclusions}
\label{sec:conclusions}
We presented a shared planning approach for a pair of personal mobility
vehicles in which each vehicle runs its own model-predictive planner, the
joint decision is expressed as utility matrices over candidate trajectory
pairs, and nothing crosses between the vehicles but observable odometry. The
behavioural parameters are fitted from a corpus of human-driven pairs, and the
geometry available to the formation is derived from the corridor rather than
fixed by hand --- offline from the map, and at runtime from what the vehicles
observe, so that an obstacle which has moved since the profile was built still
narrows the corridor the pair plans in.

In closed loop, in the two conditions the drivers were recorded in, the pair
holds conversational formation for $\resConvObsCcwOurs\%$ and
$\resConvObsCwOurs\%$ of the time among the cones (anticlockwise and
clockwise) and $\resConvFreeCcwOurs\%$ and $\resConvFreeCwOurs\%$ in a clear
aisle, against the drivers' $\resConvObsCcwDrivers\%$ and
$\resConvObsCwDrivers\%$, and $\resConvFreeCcwDrivers\%$ and
$\resConvFreeCwDrivers\%$, with no
contact with walls or cones and no vehicle left stuck. On the quantity the
formation exists to serve it therefore exceeds the humans it was fitted from,
though not by imitating them: it matches their speed, but it is more uniform
and holds a single lateral separation
where they alternated between two, and on lateral separation and stagger it
lies outside the spread among the drivers themselves. Safety rests on rules the cost cannot override: moves are executed
only if the vehicle can still brake clear, measured from the whole body, and a
vehicle that is stuck reverses along its own path and continues.

\bibliographystyle{IEEEtran}
\bibliography{IEEEabrv,mybibfile}

\end{document}